Construction Management

# Construction Material Classification on Imbalanced Datasets Using Vision Transformer Architecture (Vit)

**Maryam Soleymani** [1, *], m.soleymani.pm@gmail.com, ORCID: 0000-0003-3796-3137;
**Mahdi Bonyani** [2], m_bonyani96@ms.tabrizu.ac.ir , ORCID: 0000-0003-0922-9656;
**Hadi Mahami** [3], hadi.mahamipm@gmail.com, ORCID: 0000-0003-2426-3238;
**Farnad Nasirzadeh** [3], farnad.nasirzadeh@deakin.edu.au , ORCID: 0000-0003-0101-6322.

[1] Department of Project and Construction Management, Art University of Tehran, Tehran, Iran
[2] Department of Computer Engineering, University of Tabriz, Tabriz, Iran
[3] School of Architecture and Built Environment, Deakin University, Geelong, Australia
* Corresponding author

**Abstract:** This research proposes a reliable model for identifying different construction materials with the highest accuracy, which is exploited as an advantageous tool for a wide range of construction applications such as automated progress monitoring. In this study, a novel deep learning architecture called Vision Transformer (ViT) is used for detecting and classifying construction materials. The robustness of the employed method is assessed by utilizing different image datasets. For this purpose, the model is trained and tested on two large imbalanced datasets, namely Construction Material Library (CML) and Building Material Dataset (BMD). A third dataset is also generated by combining CML and BMD to create a more imbalanced dataset and assess the capabilities of the utilized method. The achieved results reveal an accuracy of 100 percent in evaluation metrics such as accuracy, precision, recall rate, and f1-score for each material category of three different datasets. It is believed that the suggested model accomplishes a robust tool for detecting and classifying different material types. To date, a number of studies have attempted to automatically classify a variety of building materials, which still have some errors. This research will address the mentioned shortcoming and proposes a model to detect the material type with higher accuracy. The employed model is also capable of being generalized to different datasets.

## 1. Introduction

The monitoring of construction progress is generally done in a labor-intensive process which is usually subject to error (Braun et al. 2020). An alternative solution for monitoring the project in a more accurate way is the automation of construction progress monitoring (Braun et al. 2015). For automatically monitoring construction progress or producing Building Information Models (BIM) from site images, material classification is a major step (Dimitrov and Golparvar-Fard 2014; Wang et al. 2019). Digitalized material classification not only extracts the appearance-based information for automated project monitoring and control, but also helps with the segmentation of elements for the automated generation of 3D as-built models, such as the scan-to-BIM process (Alaloul and Qureshi 2021; Dimitrov and Golparvar-Fard 2014; Son et al. 2014).

Overall, automated material classification is a beneficial tool that is utilized for a broad range of construction applications. It can improve the performance output of various activities, including defect identification, on-site material control, and progress monitoring (Alaloul and Qureshi 2021; Meroño et al. 2015; Son et al. 2014). Moreover, material classification would increase the accuracy of building element recognition significantly. Regarding its beneficial applications, it has been the subject of numerous studies in the state-of-the-art.

In the following, prior studies on material recognition and classification are reviewed, and their results are briefly summarized in Table 1. Building material detection was first proposed by Brilakis et al. (2005) to identify building elements from two-dimensional images (Brilakis et al. 2005; Han and Golparvar-Fard 2014). Then, Zhu and Brilakis presented a method of identifying concrete material regions using machine learning techniques (Zhu and Brilakis 2010). Araújo et al. introduced a methodology using application-oriented machine learning techniques for automatic classification of granite based on spectral information obtained from a spectrophotometer (Araújo et al. 2010). Kim et al. described an automatic color model–based concrete detection method through the functional analysis of three machine learning algorithms including Artificial Neural Network (ANN) model, Gaussian Mixture Modelling (GMM), and Support Vector Machine (SVM) (Kim et al. 2012). Son et al. utilized a heterogeneous ensemble classifier to increase the detection accuracy of major building materials such as concrete, steel, and wood on construction sites (Son et al. 2014). Yazdi and Sarafrazi proposed an automated system for the segmentation of concrete images into microstructures using texture analysis via an ANN classifier (Yazdi and Sarafrazi 2014). Han and Golparvar-Fard presented two methods for sampling and recognizing construction material from image-based point cloud data and inferring progress using a statistical representation from the material classification (Han and Golparvar-Fard 2014). Dimitrov and Golparvar-Fard presented a robust material classification method (a multiple one-vs.-all $x^2$ kernel SVM classifier) for semantically-rich as-built 3D modeling and construction monitoring purposes (Dimitrov and Golparvar-Fard 2014). Han and Golparvar-Fard presented a new appearance-based material classification method for operation-level monitoring of construction progress using BIM and daily construction photologs (Han and Golparvar-Fard 2015). Han et al. introduced an improved appearance-based progress monitoring method by removing occlusion; then, they used the SVM algorithm to classify materials (Han et al. 2016). Yang et al. proposed an image-based 3D modeling to recognize the materials used in a surface through images captured from different viewpoints. In this research, SVM algorithm was used (Yang et al. 2016). Rashidi et al. conducted a comparison study to evaluate the performance of different machine learning techniques including Multilayer Perceptron (MLP), Radial Basis Function (RBF), and SVM to recognize three common building material classes: Concrete, red brick, and OSB boards (Rashidi et al. 2016). Degol et al. employed 3D geometry features with state-of-the-art material classification 2D features and found that both jointly and independently modeling 3D geometry improve mean classification accuracy (Degol et al. 2016). Jiang et al. explored the application of a convolutional neural network (CNN) in classifying and identifying asphalt mixtures using the sectional images obtained from the X-ray computed tomography (CT) method (Jiang et al. 2018). Lee and Park designed and implemented a system that exhibited improved performance on the detection of reinforcing bars (Lee and Park 2019). Bunrit et al. proposed an automatic feature extraction method by CNN in transfer learning technique for construction material classification (Bunrit et al. 2019). In another study, Bunrit et al. investigated the transfer learning of GoogleNet and ResNet101 that pre-trained on the ImageNet dataset (source task) with a task specific on construction material classification (Bunrit et al. 2020). Yuan et al. introduced a terrestrial laser scanner (TLS) data-based model to classify common construction materials in which surface roughness, material reflectance, and HSV colors (Hue, Saturation, Value) are used as classification features (Yuan et al. 2020). Mahami et al. employed a new deep learning method (VGG16) to classify different building materials accurately (Mahami et al. 2020). Fernando and Marshal presented a classification methodology for excavation material identification utilizing only proprioceptive force data acquired from an autonomous digging system and three machine learning algorithms including KNN (K Nearest Neighbor), ANN, and k-means (Fernando and Marshall 2020). Davis et al. designed and described a deep CNN to identify seven typical Construction and Demolition Waste (C&DW) classifications using digital images of waste deposited in a construction site bin (Davis et al. 2021). Alaloul and Qureshi used an ANN model to classify some construction materials (Alaloul and Qureshi 2021).

Although there are several studies conducted in the area of material classification, the literature review reveals that there are some shortcomings that need to be addressed. Some of the previous studies were restricted to just one specific material (Araújo et al. 2010; Fernando and Marshall 2020; Jiang et al. 2018; Kim et al. 2012; Lee and Park 2019; Yazdi and Sarafrazi 2014; Zhu and Brilakis 2010) or a limited number of material categories (Alaloul and Qureshi 2021; Bunrit et al. 2019; Bunrit et al. 2020;

Rashidi et al. 2016; Son et al. 2014; Yang et al. 2016). Furthermore, some studies employed tiny datasets, that is to say, they contained few images (Yazdi and Sarafrazi 2014). It must be noted that tiny datasets could lead to biased training. In fact, their insufficient data will result in the inefficient performance of the trained model for new images of the material with the same class but with a different shape or color (Davis et al. 2021). Moreover, the results may be affected due to the lack of a sufficient variety of light conditions in the images of the dataset too (Rashidi et al. 2016).

**Table 1-** The summary of related studies on building material classification.

| Ref | Classified Materials | Dataset | Techniques | Achieved Outcomes | Performance |
|---|---|---|---|---|---|
| (Zhu and Brilakis, 2010) | concrete | 114 samples (train) 167 samples (validation) | SVDD, C-SVC and ANN | Identifying concrete material regions | Precision and recall: 80% |
| (Araújo et al., 2010) | granite | 16 types | Functional SVM-PUK | Functional, flexible, portable, and the overall mixed-system approach to automatic granite identification | Error rate validation: 0.82% |
| (Kim et al., 2012) | concrete | 108 images of various surfaces | GMM-ANN-SVM | An automated color model-based concrete detection | Accuracy: 93.06% |
| (Son et al., 2014) | Concrete, Steel, Wood | 108, 91, 50 images in order of the mentioned materials | an ensemble classifier | More accuracy than an excellent single classifier(namely SVM, ANN, C4.5, NB, LR, and KNN) | Accuracy: Concrete: 92.64%, Steel: 96.70%, Wood: 92.19% |
| (Yazdi and Sarafrazi, 2014) | concrete | 31 images | ANN, SVM, KNN, Bayesian, and FLD | ANN is the best system for automatic image segmentation using texture analysis | Accuracy: 90.29% |
| (Yang et al., 2016) | Tile, Brick, Stone, Coating | 504, 637, 463,and 409 samples in order | Binary SVM | A 3D material recognition strategy | Accuracy: 95.55% |
| (Rashidi et al., 2016) | Concrete, red brick, and OSB boards: | 750 images | MLP, RBF, and SVM | SVM outperformed the other two techniques in terms of accurately detecting the material textures in images | precision: concrete and OSB board: 75-95% , red brick: 94% |
| (Jiang et al., 2018) | asphalt mixtures (air voids, mastics, and aggregates) | Images of 11 asphalt mixtures segmented into three parts | CNN model | Using the sectional images obtained from the X-ray computed tomography (CT) method to classify and identify asphalt mixtures | Precision: Aggregates: 80.9%, mastics: 72.7%, air voids: 44.8% |
| (Lee and Park, 2019) | rebar | Rebar images | RFSP and IVB | A system to improve performance of detection of reinforcing bars | Precision: 99% Recall: 98% F-measure: 98% |
| (Bunrit et al., 2019) | brick, concrete, and wood | ImageNet dataset (parts of the CML) | SVM (GoogleNet and AlexNet) | Fine-tuning scheme of Googlenet reveals the highest classification result. | Accuracy: 95.50% |
| (Bunrit et al., 2020) | | 1,200 images (train), 600 images (test) | SVM (GoogleNet & ResNet101) | Autoencoder-based representation method that can improve classification more than PCA (Principal Component Analysis) in all cases | Accuracy: 97.83% |
| (Fernando and Marshall, 2020) | rock and gravel | 86 force signals (proprioceptive force data acquired from an autonomous digging system and machine learning) | KNN, ANN, k-means | A classification methodology that utilizes only proprioceptive force data for excavation material identification | Accuracy: 90% |
| (Alaloul and Qureshi, 2021) | brick, wood, concrete block, and asphalt | 50 images | ANN | A python-based ANN model for material classification | Accuracy: 64% |
| (Han and Golparvar-Fard, 2014) | Asphalt, Brick, Cement-Granular, Cement-Smooth, Concrete-Cast, Concrete-Precast, Foliage, Form Work, Grass, Gravel, Marble, Metal-Grills, Paving, Soil-Compact, Soil-Vegetation, Soil-Loose, Soil-Mulch, Stone-Granular, Stone-Limestone, Wood, Insulation, Waterproofing Paint, Steel Beam | Construction Materials Library (CML) consisting of 23 classes and 3,000 images | SVM | Two methods for sampling and recognizing construction materials from image-based point cloud data | Accuracy: 90.8% (90×90 pixel patches) |
| (Dimitrov and Golparvar-Fard, 2014) | | | binary C-SVM | A new vision-based method for material classification, robust to dynamic changes of illumination, viewpoint, camera resolution, and scale | Accuracy: 97.1% (200×200 pixel image patches) (90.8% for 30×30 pixels) |
| (Han and Golparvar-Fard, 2015) | | CML (for training/testing) and four new datasets of incomplete and noisy point cloud models (for Validation) | C-SVM | A new appearance-based material classification method for monitoring construction progress deviations at the operational-level | Accuracy: 92.4% (100×100 pixel image patches) |
| (Han et al., 2016) | | | SVM | A method to remove occlusions, prior to performing material recognition, and finally increase accuracy | Accuracy: 91% |
| (Degol et al., 2016) | Asphalt, Brick, Cement-Granular, Cement-Smooth, Concrete-Cast, Concrete-Precast, Foliage, Form Work, Grass, Gravel, Marble, Metal-Grills, Paving, Soil-Compact, Soil-Vegetation, Soil-Loose, Soil-Mulch, Stone-Granular, Stone-Limestone, Wood | GeoMat dataset (created from images supplemented with sparse 3D points of 19 material categories) | SVM | Using 2D and 3D features to improve material recognition accuracy across multiple scales and viewpoints for both material patches and images of a large-scale construction site scene | Accuracy: 91% |
| (Yuan et al., 2020) | concrete, mortar, stone, metal, painting, wood, plaster, plastic, pottery, and ceramic | Laser scan data of ten common building materials | Ensemble classifier | An automatic classification method for common building materials based on a terrestrial laser scanner (TLS) data | Accuracy: 96.7% |
| (Mahami et al., 2020) | brick, soil, sandstorms, gravel, stone, asphalt, cement-granular, wood, clay hollow block, paving, and concrete block | Consisting of 11 classes and 1,231 images | VGG16 | A method for the material classification that is not dependent on various camera angles and positions | Accuracy: 97.35% |
| (Davis et al., 2021) | second fix timbers, shuttering/ formwork timbers, shuttering/ formwork ply and particleboards, hard plastics, soft plastic, bricks and concrete, cardboards and polystyrene | First experiment (single classification): 525 images Second experiment (classification of mixed waste images): 1758 images | VGG-16 | Identifying different types of Construction and Demolition Waste (both single and mixed classes) | Accuracy: 94% |

Additionally, the accuracy rates resulted from previous studies whose datasets were more acceptable in terms of the number of images and classes of materials were in the range of 90.8% to 97.35% (Degol et al. 2016; Dimitrov and Golparvar-Fard 2014; Han and Golparvar-Fard 2014; Han and Golparvar-Fard 2015; Han et al. 2016; Mahami et al. 2020; Yuan et al. 2020). One of the biggest challenges of these studies is related to some materials with similar shapes and textures. Even when the average accuracy of previous studies is high, the accuracy of identifying similar materials drops sharply. For instance, in an experiment, grass, formwork, and brick have been detected with the most accuracy rate due to their unique pattern or texture; however, soil, cement, and concrete are recognized with much error owing to their similar shapes and textures (Dimitrov and Golparvar-Fard 2014). In another study, gravel, sand, cement-granular, and asphalt have been problematic in achieving an acceptable accuracy rate (Mahami et al. 2020). This study addresses the above-mentioned shortcomings.

This research employs a novel deep learning architecture called Vision Transformer (ViT) as a reliable classifier to maximize the accuracy of construction material detection. It should be noted that the quantity, quality, and variety of the utilized data to train models have always been the critical factors for evaluating the robustness of the implemented architecture. This is fulfilled effectively in this research since all of these factors were considered through using different imbalanced datasets. To evaluate the performance of the employed method, it is trained and tested on three large imbalanced datasets, namely Construction Material Library (CML) and Building Material Dataset (BMD), used in the previous studies, as well as a new dataset created by combining them. Finally, the comparison among results is presented. The achieved results revealed an accuracy of 100 percent. This has been unprecedented to get such a result for all the material classes and parameters, namely accuracy, precision, recall rate, and f1-score. The employed ViT classifier presents a new and promising outlook for automatic material classification.

In summary, the innovations and scientific contributions of this paper to the body of knowledge are as described below:

- The presented model is able to learn from imbalanced datasets with various amount of data within different classes and achieve 100% accuracy in all classes.
- Additionally, for the first time, the model’s generalizability was tested by two different datasets, one for training and the other for testing. Despite the high diversity within the classes, the model was still able to identify subclasses within their class correctly with great accuracy, which is a challenging situation in classification. As such, the model is capable of being generalized to different datasets and locations.

In the following section, the datasets are introduced. Then, the methodology and implementation details are described in Section 3. The next section provides the results and the validation process for the accuracy and robustness of the model. Lastly, conclusions are drawn in Section 5.

## 2. Dataset

In this study, two datasets used by previous studies were utilized for training and evaluating Vision transformer (ViT) model.

The first dataset created by Mahami et al. (Mahami et al. 2020), henceforth referred to as Building Material Dataset abbreviated to BMD, is consisting of 1231 images. The images have a resolution of 4608×3456 pixels, captured from different distances, viewpoints, and light conditions. The dataset deals with 11 classes, applied to identify different materials such as asphalt, brick, wood, etc. (Figure 1). Also, the distribution of images in each material class was unequal. Lack of uniformity in the distribution of data within the classes of a dataset is interpreted as an imbalanced dataset (Japkowicz and Stephen 2002), shown better in Figure 1.

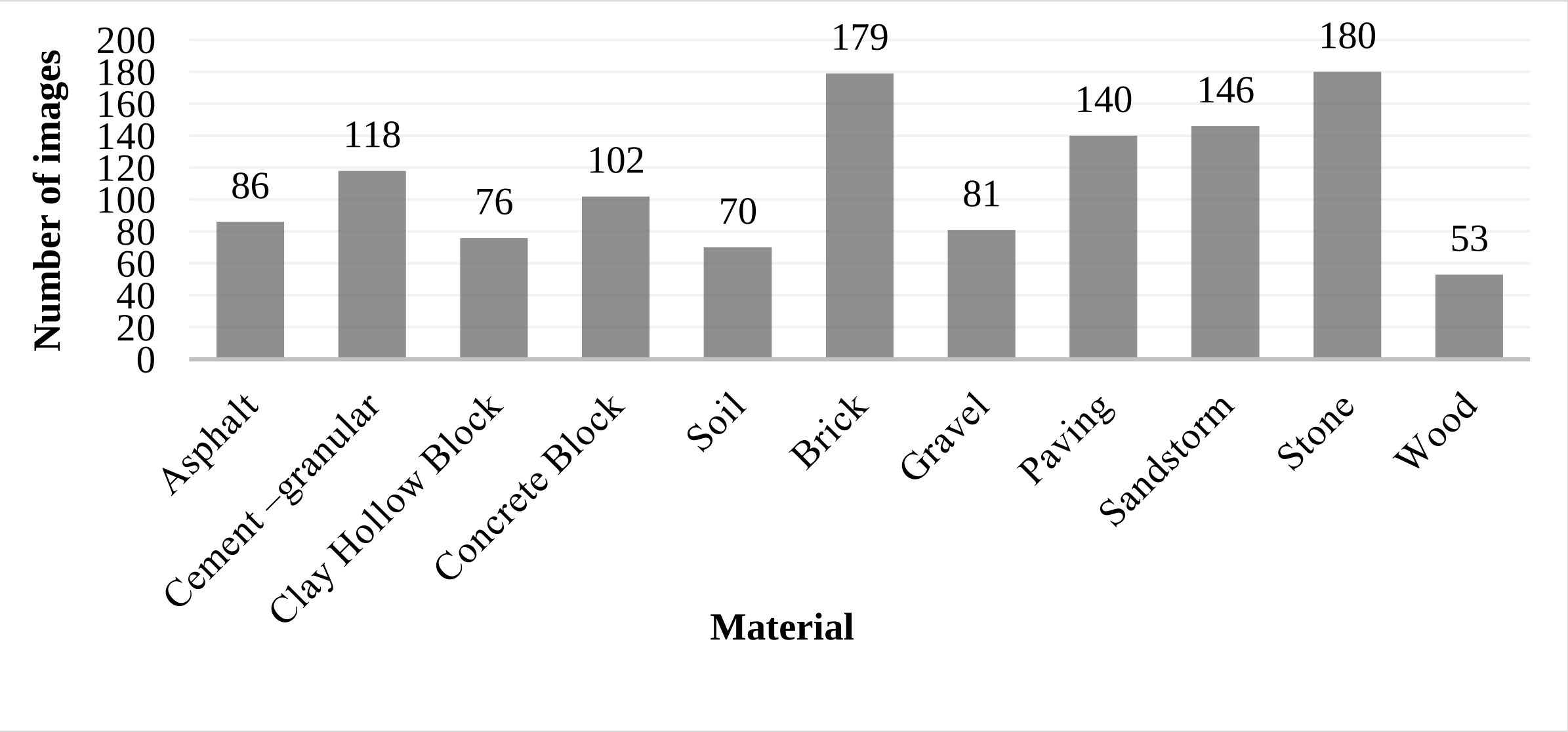

**Figure 1.** The distribution of images of the BMD dataset among different classes.

The second dataset is the Construction Materials Library (CML) including a wide range of materials and intra-class variability, which was created by Dimitrov and Golparvar-Fard (2014) and improved by Han and Golparvar-Fard (2015), used in several articles(Dimitrov and Golparvar-Fard 2014; Han and Golparvar-Fard 2014; Han and Golparvar-Fard 2015; Han et al. 2016) for material recognition purposes. Moreover, there are previous studies that have employed some parts of CML (Bunrit et al. 2019; Bunrit et al. 2020). The total number of images in the CML is 3266 which have a fixed resolution of 200×200 pixels, consisting of 20 classes. The class distribution was imbalanced, as depicted in Figure 2.

In addition, from the combination of these datasets, a new composite dataset was created, including 4497 images and 24 classes of materials, as shown in Figure 3. The important characteristic of this dataset is the variety in size and quality of images and also the unequal class distribution causing an imbalanced dataset. This leads to a more challenging project as learning from such data is a common problem in classification. In fact, the classification of data with imbalanced class distribution has encountered a significant drawback of the attainable performance by most standard classifier learning algorithms (YANMIN SUN). Moreover, highly imbalanced data poses added difficulty, as most learners will exhibit bias towards the majority class, and in extreme cases, may ignore the minority class altogether (Johnson and Khoshgoftaar 2019).

There are samples of images in material classes of BMD and CML datasets represented in Figures 4 and 5, respectively. As can be seen, some materials have a unique pattern, such as brick and formwork that makes classification less challenging. While others are of similar appearance properties, like sandstorm, gravel, asphalt, cement-granular, and stone-granular, which leads to a more problematic classification. However, we overcame this challenge.

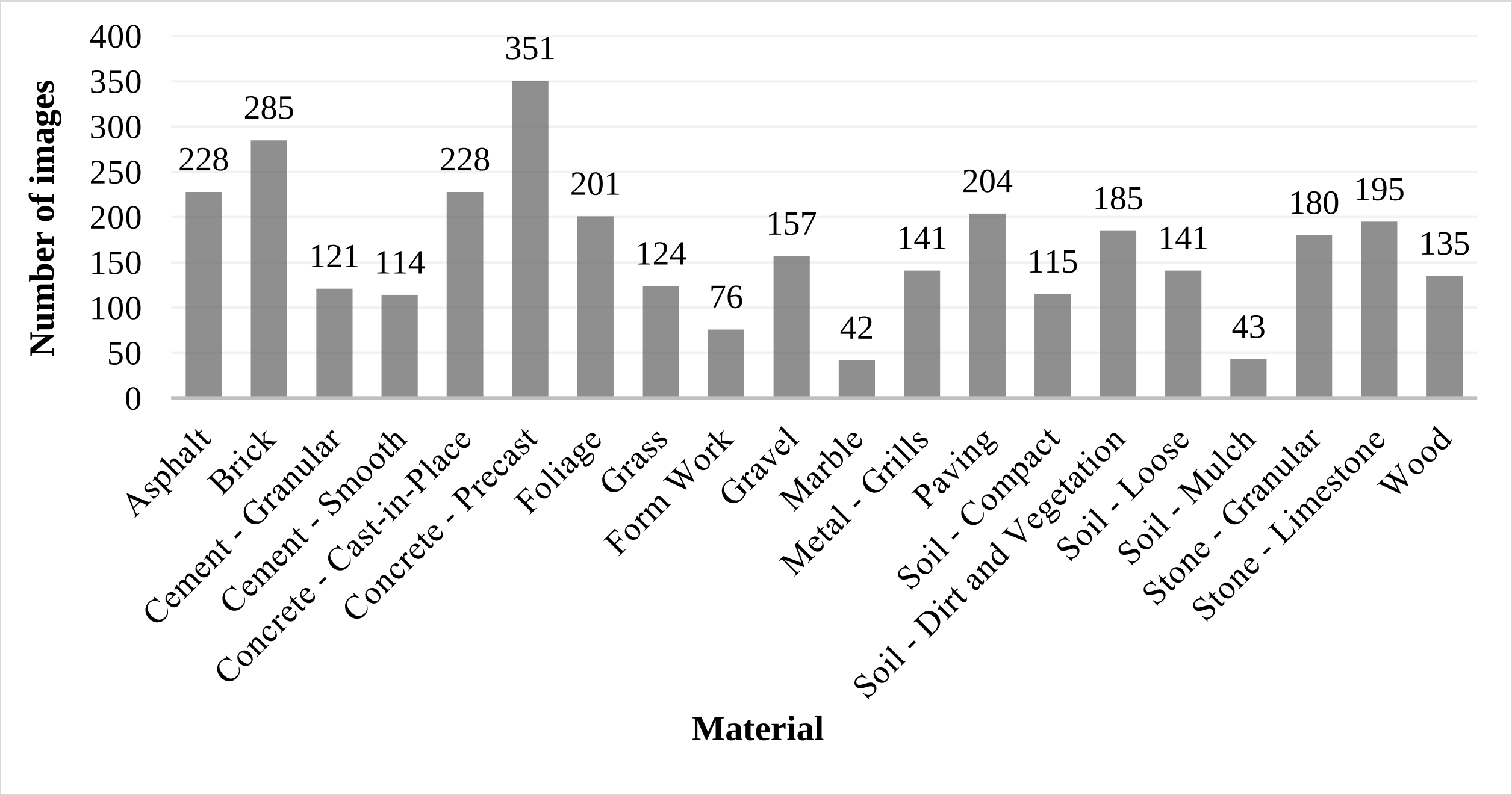

**Figure 2.** The distribution of images of the CML dataset among classes.

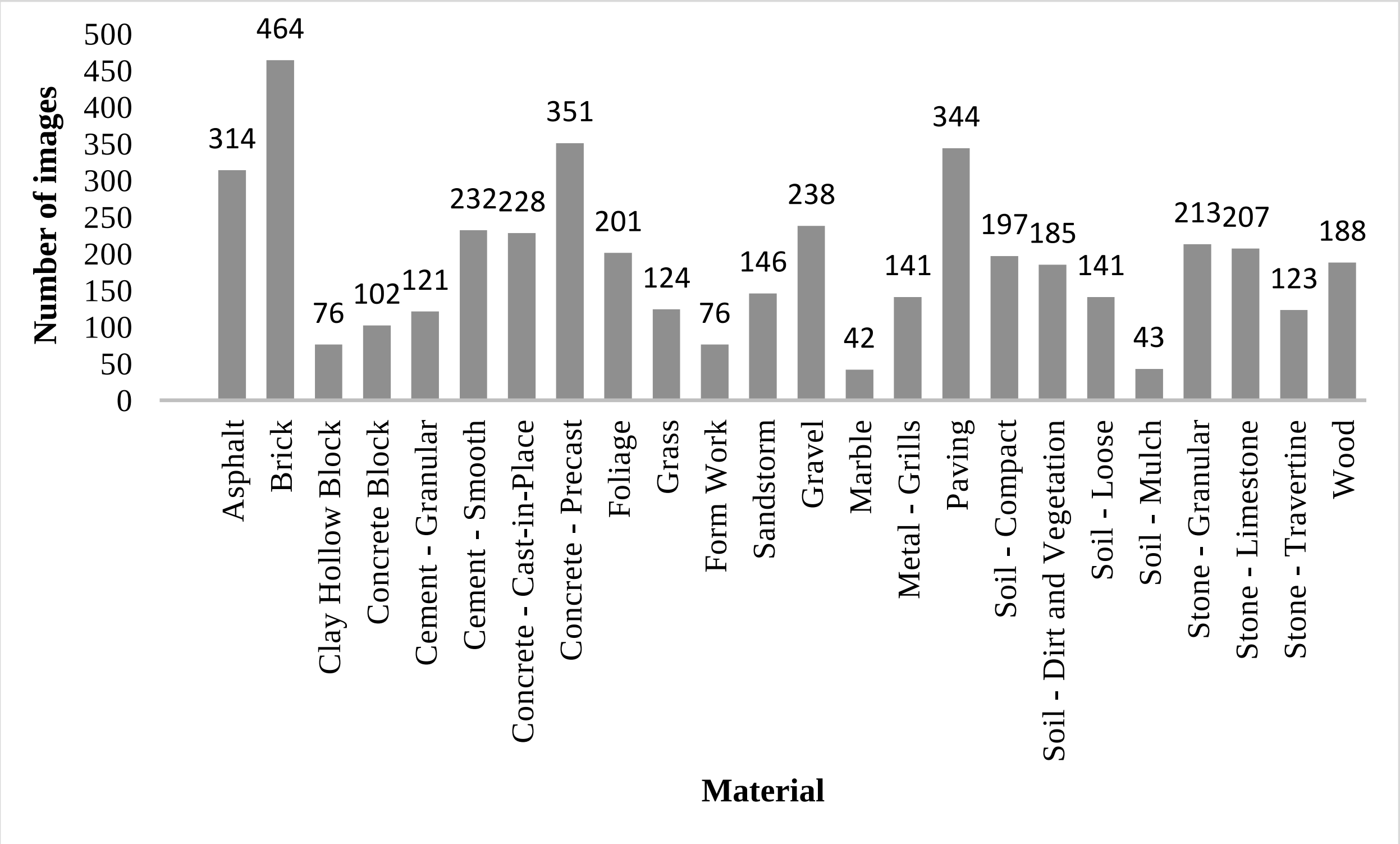

**Figure 3.** The distribution of images of the combined dataset among classes.

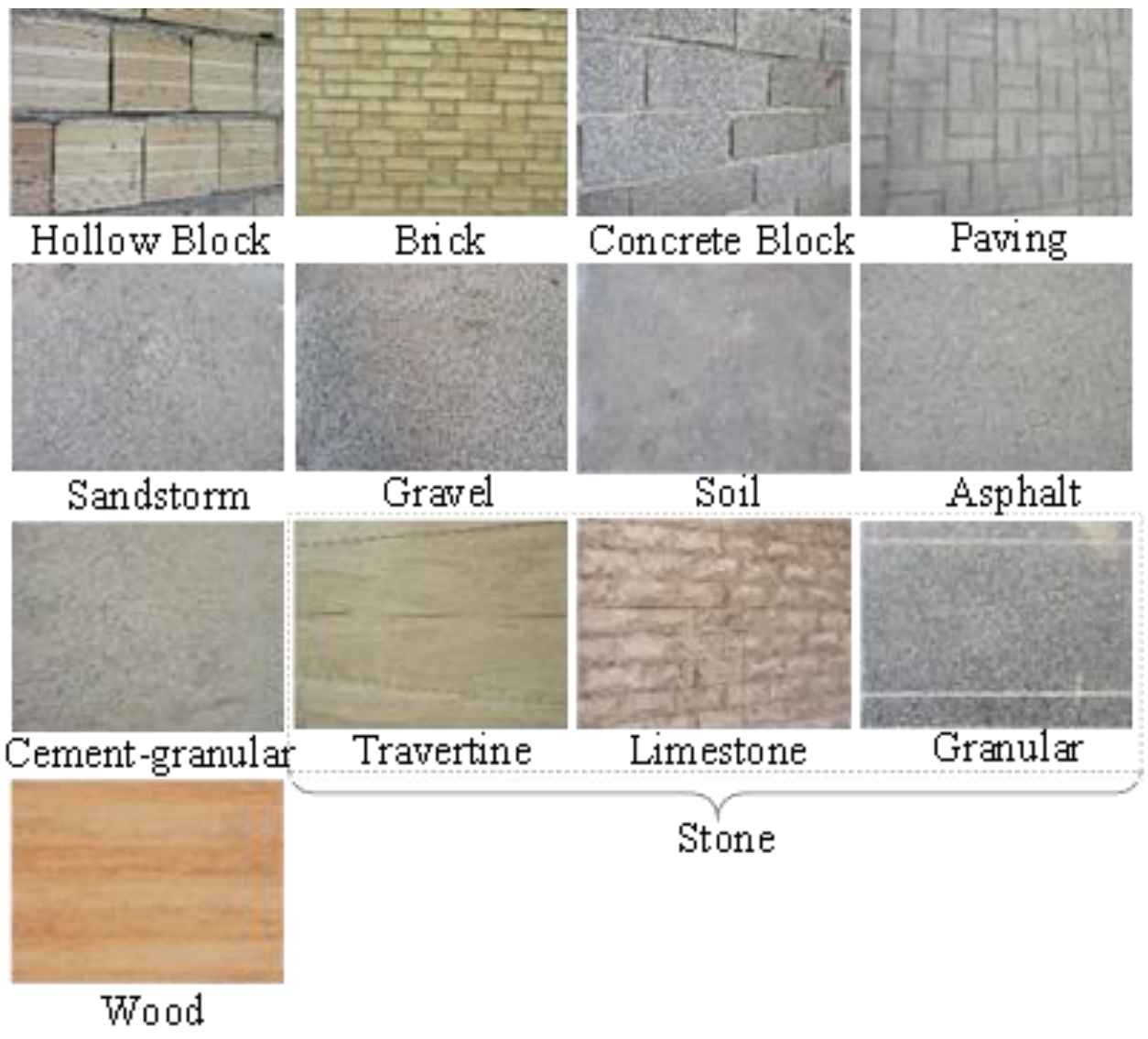

**Figure 4.** The categories of materials in the BMD dataset.

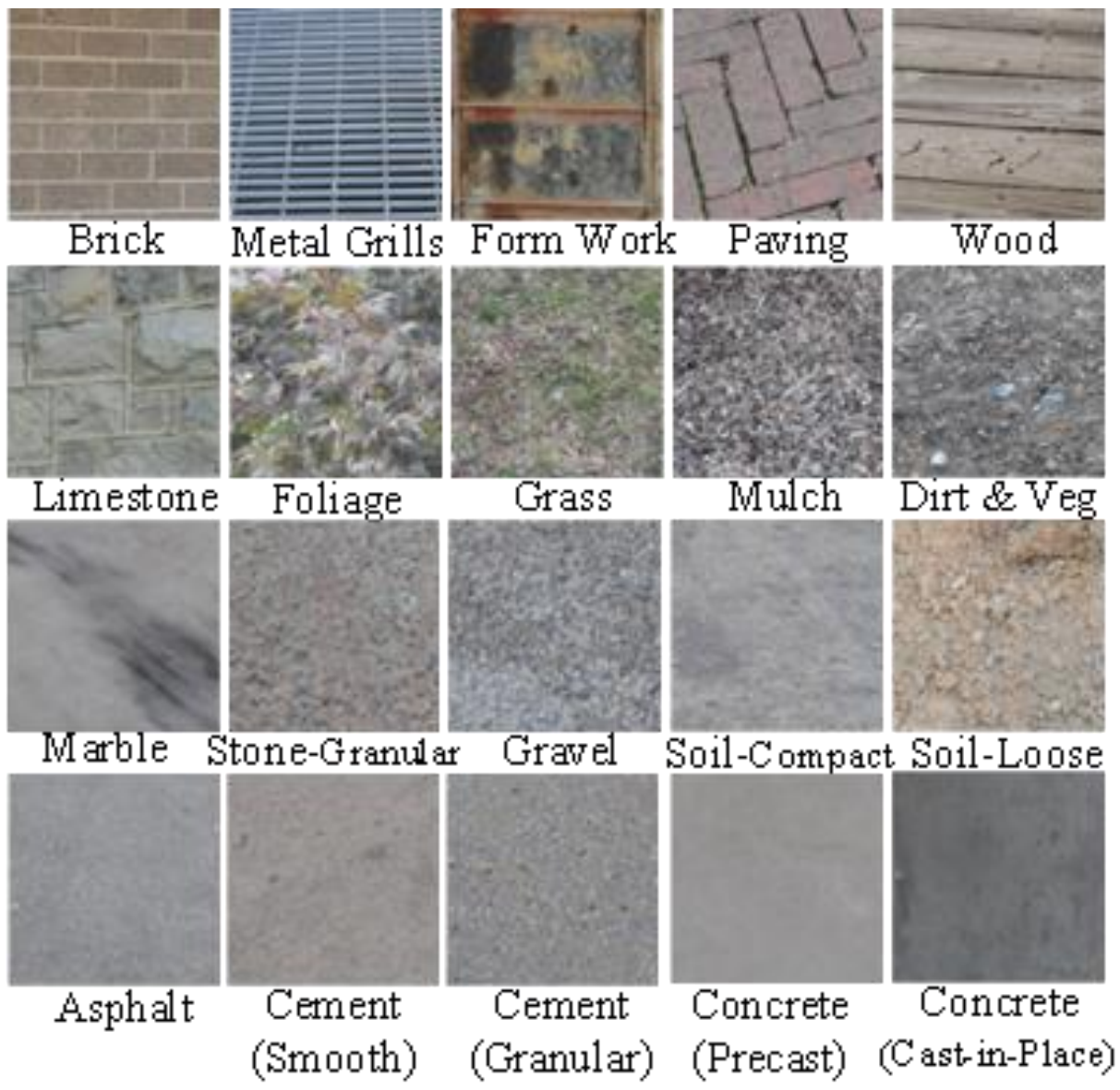

**Figure 5.** The categories of materials in the CML dataset.

## 3. Methodology

### 3-1. Architecture

An important part of deep learning is the use of Convolutional Neural Networks (CNNs). Many studies in recent years have used famous network architectures like VGG-16 (Simonyan and Zisserman 2014), DenseNet (Huang et al. 2017), GoogleNet (Szegedy et al. 2016; Szegedy et al. 2014), and EfficientNet (Tan and Le 2019). A typical convolutional network consists of convolutional layers and pooling layers, and a few fully connected layers. Several studies have suggested that the accuracy can be increased by utilizing a deeper network. With the spatial stream ConvNet, performance and computing efficiency have been balanced (Masood et al. 2020) by using the network configuration of the VGG-16 (Simonyan and Zisserman 2014), which was proven to extract the features of the images layer by layer and worked for classifying the images. In spite of the simplicity and efficiency of the VGG-16 configuration, it is a large number of parameters on fully connected layers that pose a major problem (Sermanet et al. 2013; Simonyan and Zisserman 2014). However, CNNs come with shortcomings. This is because convolution operates on a fixed-size window. Therefore, the model cannot recognize long-range spatial relationships between different parts of the images or between individual images. In addition, there are other challenges such as layer pooling which leads to information loss, as well as translation invariance which prevents models from dynamically adapting to changes in the input (Gu et al. 2018).

Attention-based architectures like Transformers (Vaswani et al. 2017) have been developed into the Natural Language Processing (NLP) domain to optimize various language-related tasks more effectively, such as translations and text classification. Transformers have resulted in significant performance gains, which have caused a great deal of interest in the computer vision community to apply similar self-attentional models to vision tasks. Several computer vision studies have been inspired by the success of self-attention in NLP that incorporate self-attention in CNN-like architectures (Carion et al. 2020; Wang et al. 2017). A few works have used the self-attention mechanism (Ramachandran et al. 2019; Wang et al. 2020) as an alternative to the convolution mechanism. Despite their efficiency, these models haven't been able to scale on modern hardware accelerators because they require specialized attention patterns (Wang et al. 2017).

The Vision Transformer (ViT) architecture was introduced by Dosovitskiy et al. (2020) (Dosovitskiy et al. 2020) for the first time. An attempt is made to design a deep neural network algorithm that does not use convolution operations over large datasets. For this purpose, ViT employs the original transformer which was developed in (Vaswani et al. 2017) for NLP tasks with a few changes. The general concept of Vit is to divide the image into patches, then flatten and project them into a D-dimensional embedding space obtaining the so-called patch embedding. In the viewpoint of NLP, a new network will be trained in a supervised learning manner by first treating image patches as tokens (words) and then using them as inputs. Next, it adds positional embedding i.e., a set of learnable vectors allowing the model to retain positional information and concatenate a (learnable) class token, then let the Transformer encoder do its magic. As opposed to CNNs which function on spatial domain-specific information, transformers work on vectors and require much more and larger datasets to discover knowledge. Finally, a classification head is applied to the class token to obtain the model's logits. The model's performance was acceptable when trained on ImageNet (1M images), great when pre-trained on ImageNet-21k (14M images), and state-of-the-art when pre-trained on Google's internal JFT-300M dataset (300M images) (Dosovitskiy et al. 2020).

The main architecture of the transformer encoder comes from Vaswani et al. (Vaswani et al. 2017). A multiheaded self-attention mechanism is layered on top of different encoder blocks for each block. Prior to being fed into the multi-head self-attention, and also prior to the MLP blocks, layer normalization is applied to the embedded patches. An encoder's general mechanism is illustrated in Figures 6 and 7 providing a general explanation of how encoders, multi-head attention, and scaled-dot product attention work (Dosovitskiy et al. 2020).

With previous training on large datasets, and experience in a wide range of mid-sized or small image recognition benchmarks, ViT achieves great results in comparison with state-of-the-art CNNs while it needs considerably fewer computational resources to train (Dosovitskiy et al. 2020).

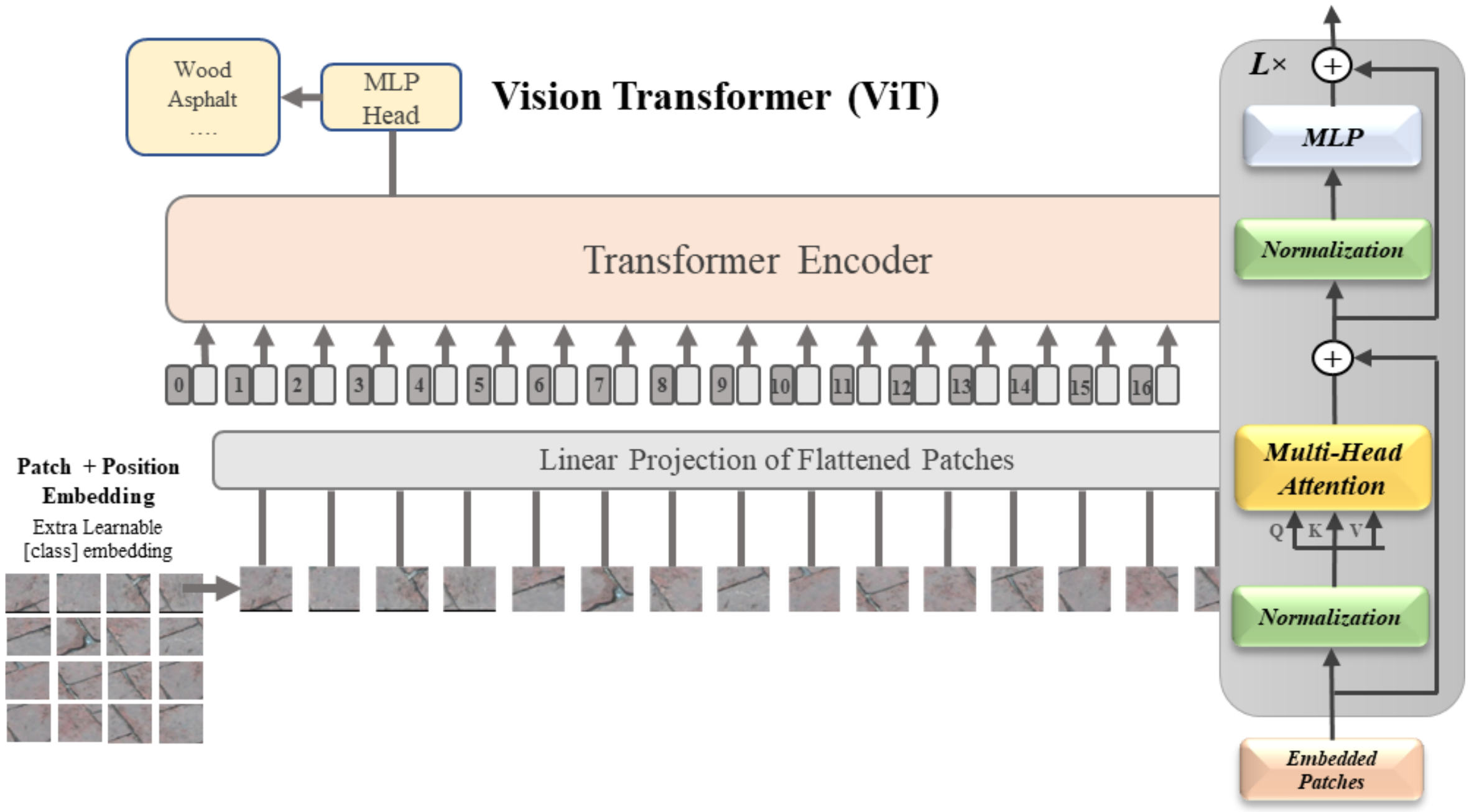

**Figure 6.** Schematic illustration of ViT model (Dosovitskiy *et al.*, 2020).

All in all, how CNNs have been applied to computer vision tasks has been largely driven by advances in GPU technology and modern computational power. In a wide range of applications, consisting of challenging image detection tasks on ImageNet, several architectures have achieved state-of-the-art performance (Krizhevsky et al. 2012). Although CNNs have many advantages, they have some limitations, namely locality, translation invariance, and max pooling (Wang et al. 2017). Vision transformers are able to resolve CNN challenges by addressing an image first as a sequence of patches, and then decoding them as a standard Transformer encoding, in contrast to many earlier tools inspired by NLP based on image analysis. This strategy is confirmed that can perform as well as CNN on natural image classification tasks when it is trained on large datasets (Dosovitskiy et al. 2020).

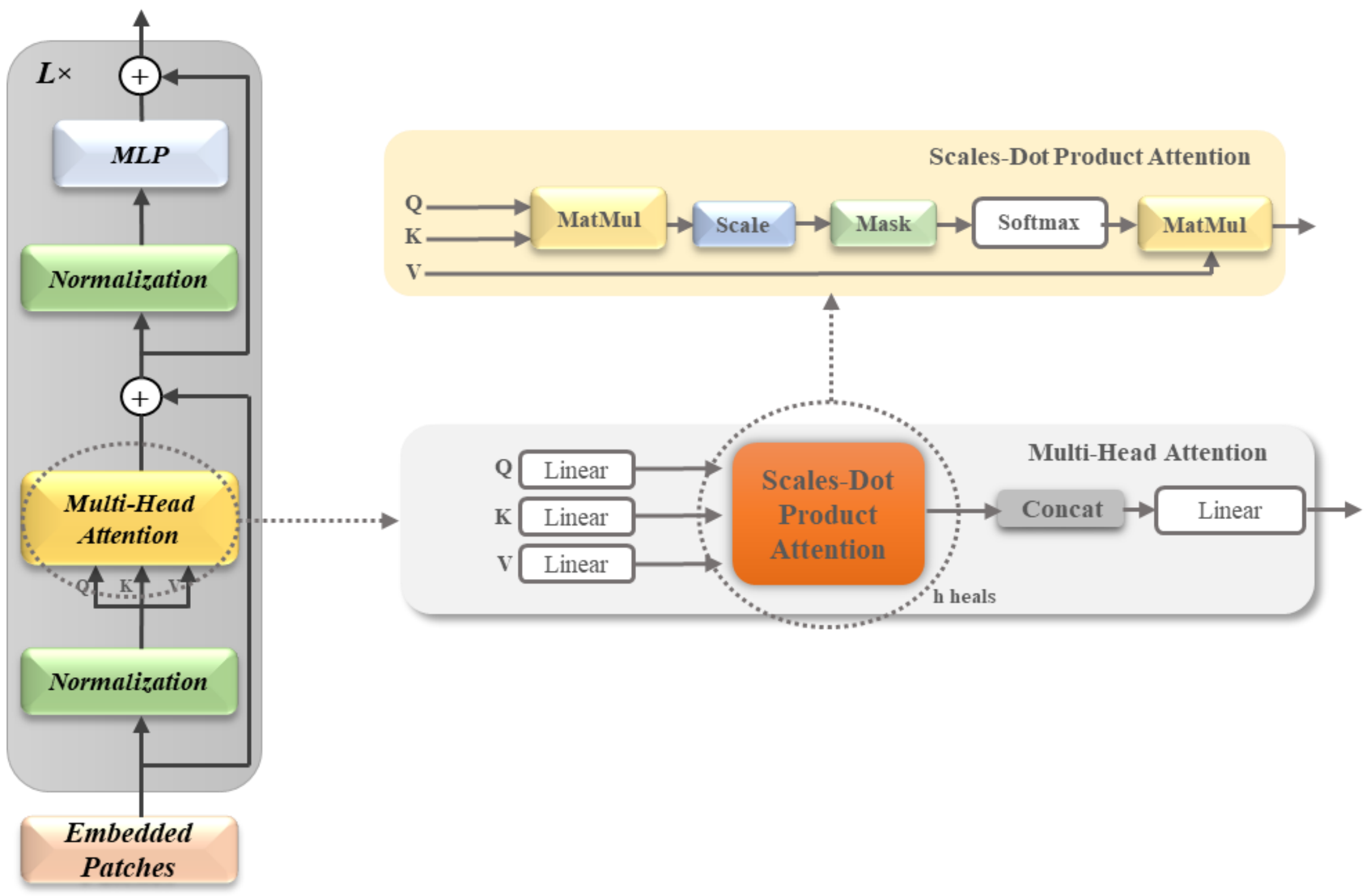

**Figure 7**. Multi-head attention mechanism (Vaswani *et al*., 2017).

### 3-2. Pre-Processing and Data Augmentation

Due to the different resolutions of the images, the only pre-processing done was to resize the images to 224×224 pixels. However, the utilized data augmentation included Flip-Left-to-Right (FlipLR), Flip-Up-to-Down (FlipUD), translate, random-crop, and Randaug (Cubuk et al. 2020), shown in Figure 8. These were done during training, then the images were normalized. Translation involved moving the image in X or Y direction (or both). In this work, the image was moved by 16 pixels in both directions. Likewise, the Randaug method with n=2 and m=7 involved mixing randomly generated augmentations (equalize, autocontrast, posterize, equalize, solarize, brightness, and sharpness).

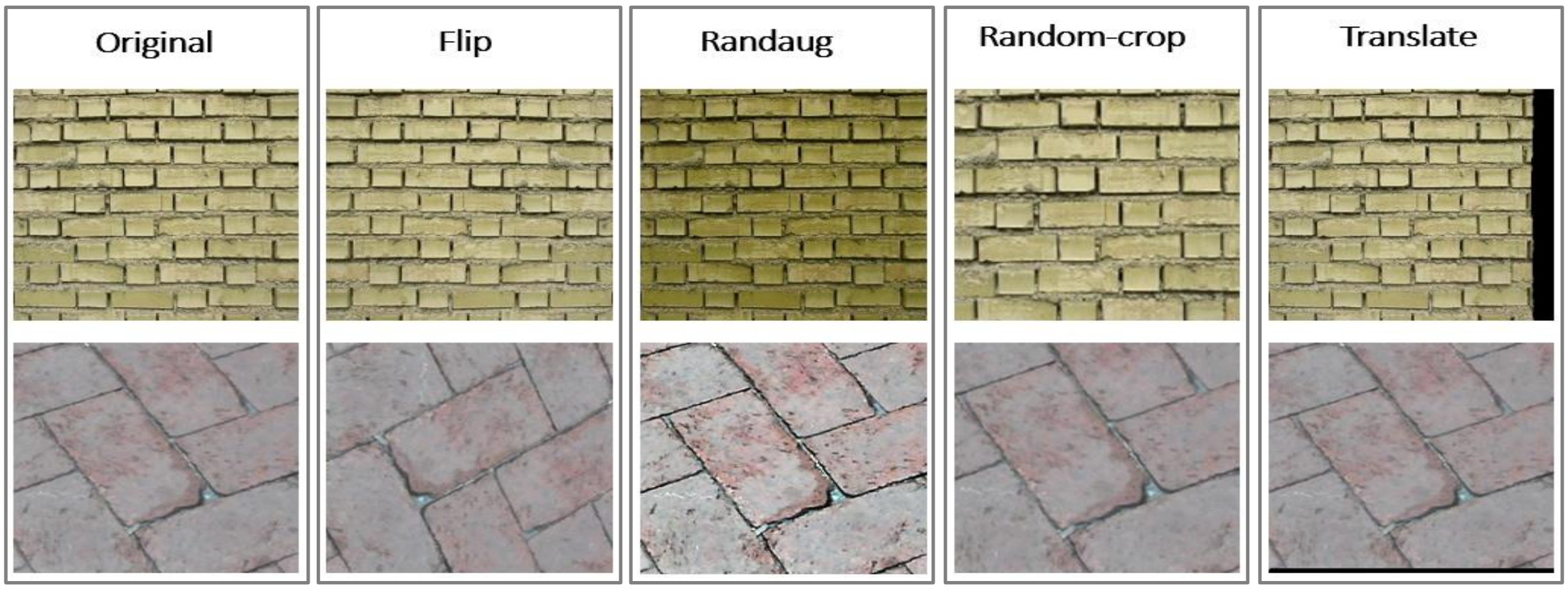

**Figure 8.** Some samples of applied augmentations (randaug, random-crop, translate). Each row illustrates a sample image and the applied augmentation is shown on top of each column.

We used the basic Vision Transformer model with a size of 16×16 patches to extract the features and the weights of the pre-trained model used on the original images. The model is 25 epochs with a learning rate of 0.0003, the Adam (Kingma and Ba 2014) optimization function, batch size 8, and the cross-entropy loss function, which have been employed for training. We also took advantage of Google Colab Graphics processing Tesla K8 Unit for training.

### 3-4. Network Evaluation

We used four standard performance metrics, described in equations (1) to (4): accuracy, precision, recall rate, F1-score as well as the confusion matrix. The elements to calculate the mentioned metrics are True Positive (TP), True Negative (TN), False Positive (FP), and False Negative (FN) that are defined as follows to evaluate the performance of the model:

- True Positive (TP) = correct material class prediction
- False Positive (FP) = incorrect material class prediction
- True Negative (TN) = correctly predicts the negative material class
- False Negative (FN) = incorrectly predicts the negative material class

**(1)** $$Precision = TP/(TP + FP)$$

**(2)** $$Recall = TP/(TP + FN)$$

**(3)** $$F1 - score = (2 * Precision * Recall)/(Precision + Recall)$$

**(4)** $$Accuracy = (\mathrm{TP} + \mathrm{TN})/(\mathrm{TP} + \mathrm{TN} + \mathrm{FP} + \mathrm{FN})$$

## 4. Results and Discussion

In this section, we first evaluate the model separately on the datasets used by the various segmentation modes that are intended for the data, and finally, we examine the reliability of these partitions by the 5-fold cross-validation method. In the following, we create a new and more challenging situation for the model by combining two datasets including BMD and CML. In this case, the model is evaluated by different data divisions.

### 4-1. Evaluating the BMD and the CML Datasets

In order to evaluate the model more accurately, we assessed the datasets in the most different divisions of the validation test as well as the 5-fold cross-validation mode, which is shown in Tables 2 and 3. In all these cases, except for one case on two datasets, we obtained an accuracy of 100%. The imbalance in the amount of data of datasets in test phase can be seen in fusion matrixes in Figures 9 and 10. Finally, the performance of the employed method was compared with the existing state-of-the-art methods listed in Table 4.

**Table 2.** Results obtained in different modes of data division for training and testing the BMD dataset (in terms of percentage).

| Train | Validation | Test | Accuracy Validation | Accuracy Test |
|---|---|---|---|---|
| 85 | - | 15 | - | 100 |
| 70 | - | 30 | - | 100 |
| 70 | 15 | 15 | 100 | 100 |
| 60 | 20 | 20 | 100 | 100 |
| 60 | 10 | 30 | 100 | 100 |

**Table 3**. Results obtained in different modes of data division for training and testing the CML dataset (in terms of percentage).

| Train | Validation | Test | Accuracy Validation | Accuracy Test |
|---|---|---|---|---|
| 85 | - | 15 | - | 100 |
| 70 | - | 30 | - | 100 |
| 70 | 15 | 15 | 100 | 98.21 |
| 60 | 20 | 20 | 100 | 100 |
| 60 | 10 | 30 | 100 | 100 |

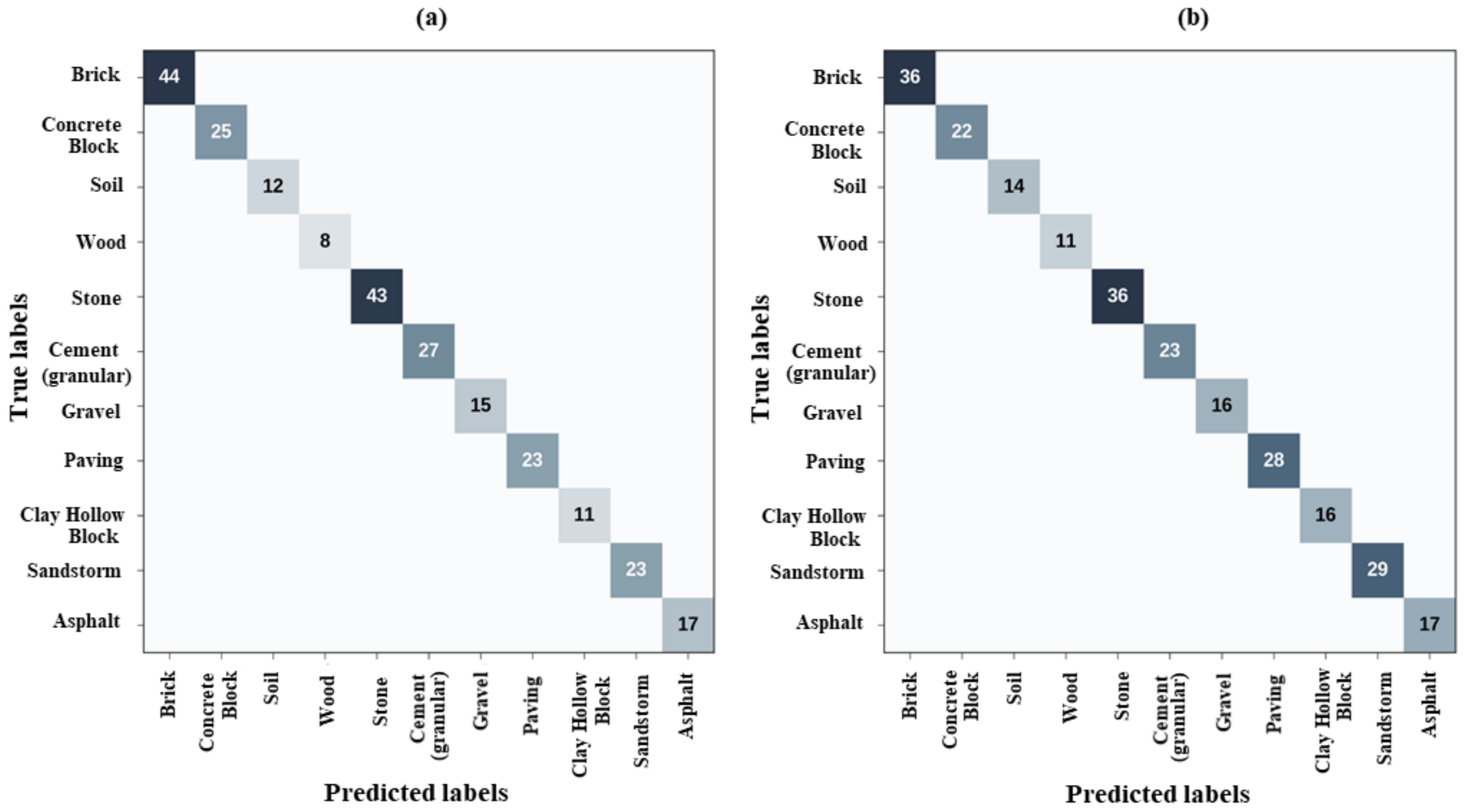

**Figure 9.** Fusion matrix on the BMD dataset for: (a) data division (b) 5-fold cross-validation.

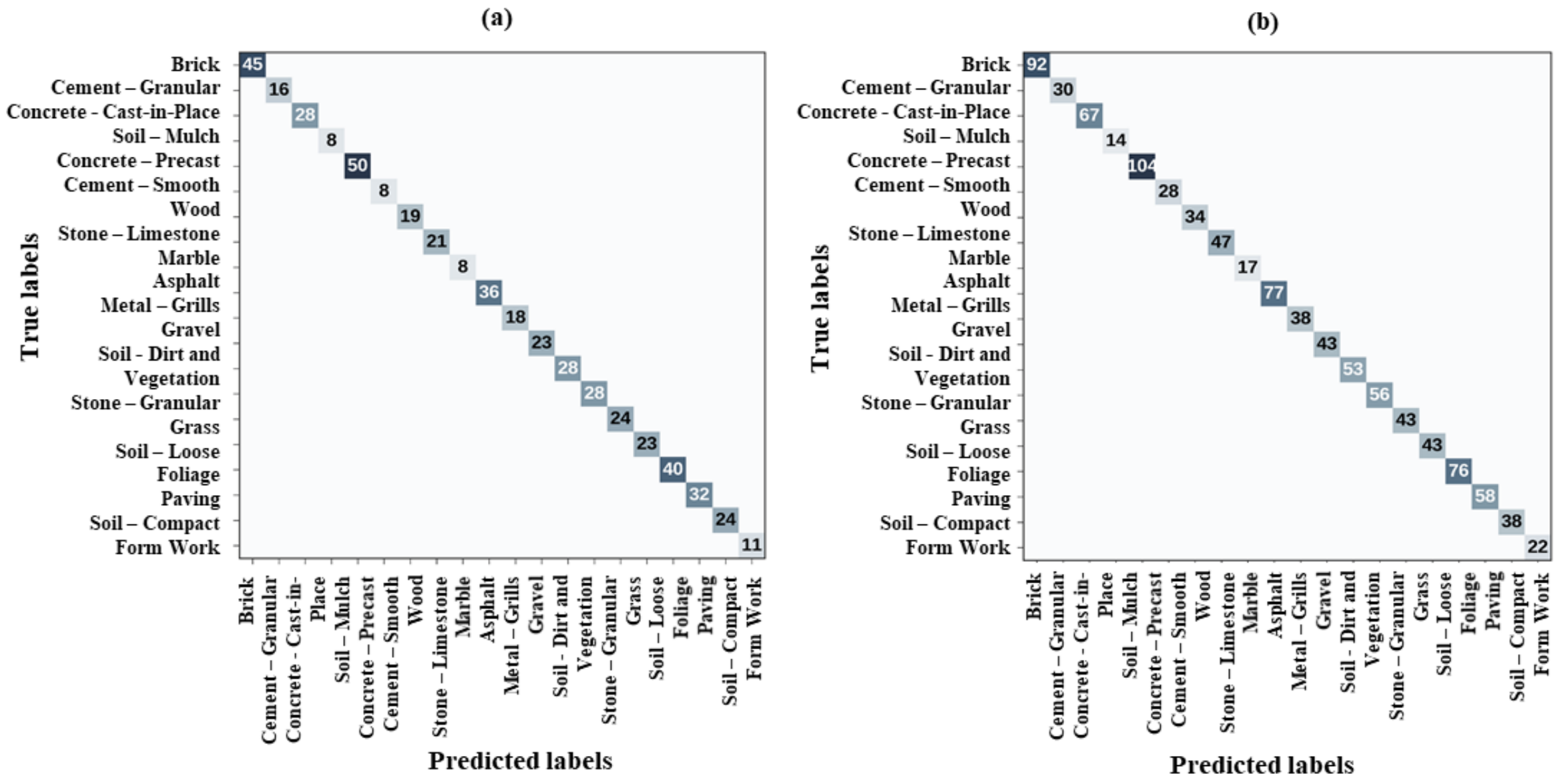

**Figure 10.** Fusion matrix on the CML dataset for: (a) data division (b) 5-fold cross-validation.

**Table 4.** Comparison of the results of utilizing ViT on the BMD and the CML datasets with the past studies.

| Reference | Dataset | accuracy | precision | recall | f1-score |
|---|---|---|---|---|---|
| (Mahami et al., 2020) | BMD | 97.35 | - | - | - |
| **This paper** | | **100** | **100** | **100** | **100** |
| (Dimitrov and Golparvar-Fard, 2014) | CML | 97.1 | - | - | - |
| (Han and Golparvar-Fard, 2014) | | 90.8 | - | - | - |
| (Han and Golparvar-Fard, 2015) | | 92.4 | - | - | - |
| (Han et al., 2016) | | 91 | - | - | - |
| (Bunrit et al., 2019) | | 95.5 | - | - | - |
| (Bunrit et al., 2020) | | 97.83 | - | - | - |
| **This paper** | | **100** | **100** | **100** | **100** |

Both datasets used do not have a default partition to evaluate the model used. Therefore, in order to make a fairer comparison with related work, we evaluated both datasets in different modes of data segmentation (Tables 2 and 3). According to the obtained results, our model was able to achieve 100% accuracy in all cases except one (Table 3). Moreover, for better evaluation, we used the 5-fold cross-validation method (Table 5) to be able to have a more accurate assessment of the model's behavior relative to different parts of the data. By doing this, we can evaluate the generalizability of the model for different partitions of data and in different conditions.

**Table 5.** The percentage of correctly recognized materials using 5-fold cross-validation on accuracy.

| Dataset | Fold 1 | Fold 2 | Fold 3 | Fold 4 | Fold 5 | Mean ±std |
|---|---|---|---|---|---|---|
| BMD | 100 | 100 | 100 | 100 | 100 | 100 ± **0.0** |
| CML | 100 | 100 | 100 | 100 | 100 | 100 ± **0.0** |

### 4-2. Combining Two Datasets

In this section, the two datasets including BMD and CML are combined leading to more subclasses and variety. This is performed to test the model's performance in a more complex way than a separate evaluation for both datasets. In this case, the number of classes for material detection increased to 24. Moreover, the model will have more challenges to achieve 100% accuracy due to the difference in resolution of the two datasets, the way of photography, and the type of texture of different materials. In this case, the model is evaluated in different modes of dataset division similar to the individual evaluation of two datasets.  In order to better evaluate and measure the generalizability of the model performance on different parts of the dataset, the 5-fold cross-validation method is also used.

According to the results shown in Table 6, it can be seen that the model has no over-fit on the test dataset, and the model can generalize to the test and validation dataset for different modes. In addition, the model has reached 100% accuracy on the validation and testing data in the case where 80% of the data is used for training, 10% for validation, and 10% for testing, which is the best possible result.

**Table 6.** Results obtained in different modes of data division for training and testing the combined dataset (in terms of percentage).

| Train | Validation | Test | Accuracy Validation | Accuracy Test | TTA |
|---|---|---|---|---|---|
| 85 | - | 15 | - | 98.75 | 99.70 |
| 70 | - | 30 | - | 99.70 | 100 |
| **80** | **10** | **10** | **100** | **100** | **100** |
| 70 | 15 | 15 | 98.75 | 98.75 | 100 |
| 60 | 20 | 20 | 99.10 | 99.10 | 99.70 |
| 60 | 10 | 30 | 100 | 99.40 | 99.64 |

However, in other cases where the model could not reach 100% accuracy on the test dataset, we can point to the imbalance in the amount of data in each class of two datasets. This imbalance can also be seen in the combined dataset. As such, data division will be a more complex situation. In fact, in both datasets, the images of the same classes can have very different textures such as paving or wood. Moreover, it is possible that the textures of the images are similar in different classes like soil or concrete. These two can cause the model to be misleading. It creates the distribution of each material in the dataset more challenging which makes the model unable to achieve 100% accuracy. As a result, we evaluated the model with a Test Time Augment (TTA) method (Shanmugam et al. 2021). In this algorithm, the original image and images augmented by training augment are separately given to the model and the max vote is selected as the final result. Consequently, the accuracy of the model increased in all used divisions. The effect of different data division modes used for training and validation on the final result indicates that the used algorithm is powerful enough for the various amount of data within classes of imbalanced datasets.

The 5-fold cross-validation method was also used to evaluate the model in the combination of two datasets. The model achieved 100% accuracy on folds one and two and 97.32%, 99.10%, and 99.10% on folds 3 to 5. However, according to the results of the 5-fold cross-validation method, it can be concluded that the obtained results represented in Table 6 are reliable and are not affected by the random division of data.

### 4-3. Generalizability test

To evaluate whether the presented algorithm could be generalized, one of the datasets was used for training and the other for testing, then vice versa. The datasets were produced in different countries and have some common material classes with different appearance. Figure 11 presents different paving materials that are used in in different countries as an example. According to Table 7, when the BMD dataset was used for testing, Test Time Augment (TTA) became 97.61 which is still higher than the past studies and when the CML dataset was used for testing, TTA became 94.79. Actually, the results indicate that though the accuracy did not become exactly 100 percent, this model has a high generalizability. Also, TTA results show that various data augmentations employed were useful and have improved the accuracy.

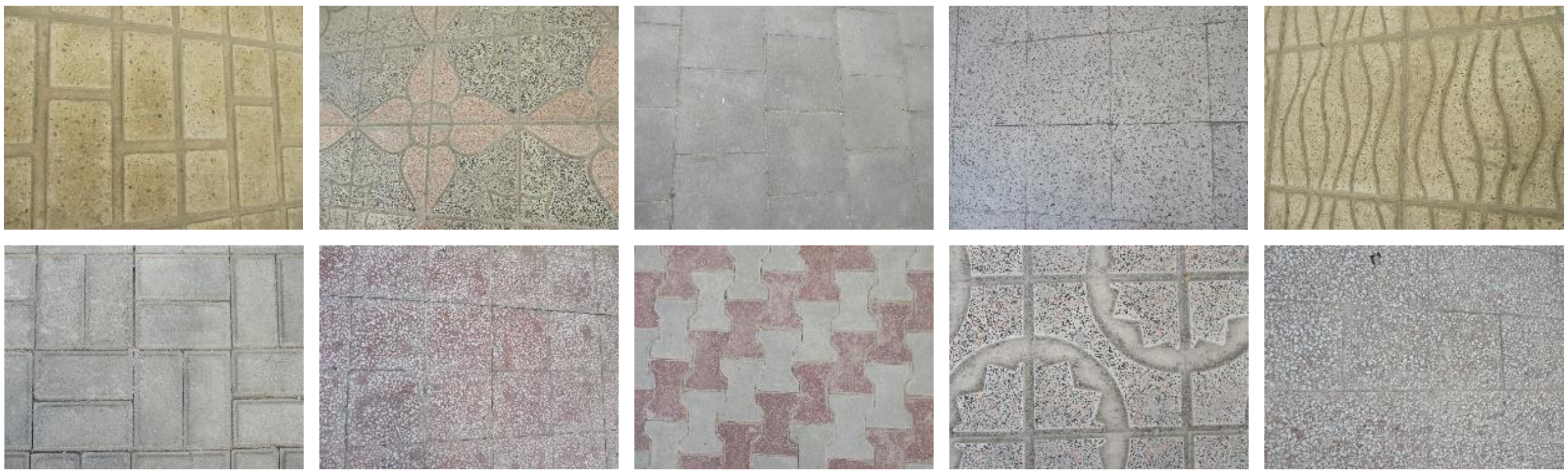

**Figure 11.** Different appearance characteristics (color, pattern, texture, etc.) in paving materials in different countries.

**Table 7.** Results obtained from training one dataset and testing the other dataset.

| Train Data | Test Data | Accuracy Test | TTA |
|---|---|---|---|
| CML | **BMD** | 96.23 | **97.61** |
| BMD | **CML** | 93.40 | **94.79** |

Our model used two different datasets for training and testing. Therefore, it is believed our model is the first that can be generalized to different image datasets in reality. In comparison to previous studies (Table 4) which utilized just one dataset for both training and testing, the results of this study are still better except for one case (Dimitrov and Golparvar-Fard 2014) whose accuracy rate was 97.1 using the CML dataset, which is 2.31% higher than our result. Also, this comparison is regardless of Bunrit et al. 2019 and 2020 that worked on just three materials and did not face serious challenges.

The reasons of the minor error we arrived can be summarized as below:

- The two datasets have many significant differences in size and image resolutions, which were described in the dataset section completely. Such differences in quality and quantity make the image classification considerably challenging. But the used model met the challenges successfully.
- The BMD dataset was created in Iran and the CML was produced in the USA. In spite of different locations which cause high variety in the texture, color and pattern within some material classes and also different light conditions, the achieved accuracy rate is still noticeable.

Despite encountered challenges, the result of generalizability test shows that the model could be customized in different situations by importing a small part of the new dataset to the dataset which is used for training the model. This would lead to reducing or removing these small errors too.

## 5. Conclusion

Automated progress monitoring provides an accurate and timely assessment of project progress. One of the controversial and essential issues in automated progress monitoring is material classification.

Although previous studies have used machine learning for automated material recognition, they still have some limitations. One of the biggest challenges is related to their high error in detecting the materials with similar shapes and textures. Also, the generalizability to other datasets was not tested. To address these shortcomings, this research utilized a novel deep learning architecture called Vision Transformer (ViT). To evaluate the performance of the employed model, it was implemented on three various imbalanced datasets in different data segmentation modes. Moreover, various data augmentations were used in order to solve the problems of data shortage, overfitting, and imbalanced classes so that the model's generalizability is kept as high. Considering evaluation metrics such as accuracy, precision, recall rate, and f1-score, ViT model outperforms the other state-of-the-art works. Not only did the ViT model achieve a rate of 100% in all four parameters, but also it accomplished this accuracy rate in all material categories in three different datasets. Besides, this model revealed a high generalizability to use different datasets for training and testing process. This shows the merits of the employed method over the past studies. It is believed that the utilized method provides a powerful tool to recognize and classify various material types.

## Data Availability Statements

All data, models, and code that support the findings of this study are available from the corresponding author upon reasonable request.

## Statements and Declarations

The authors declare that they have no known competing financial interests or personal relationships that could have appeared to influence the work reported in this paper. Moreover, this research did not receive any specific grant from funding agencies in the public, commercial, or not-for-profit sectors.

## Acknowledgement

The authors would like to thank Dr. Kevin K. Han for providing access to the CML dataset.